\documentclass[letterpaper]{article} 
\usepackage{aaai25}  
\usepackage{times}  
\usepackage{helvet}  
\usepackage{courier}  
\usepackage[hyphens]{url}  
\usepackage{graphicx} 
\usepackage{natbib}  
\usepackage{caption} 
\usepackage{algorithm}
\usepackage{algorithmic}

\usepackage{newfloat}
\usepackage{listings}
\DeclareCaptionStyle{ruled}{labelfont=normalfont,labelsep=colon,strut=off} 
\floatstyle{ruled}
\newfloat{listing}{tb}{lst}{}
\floatname{listing}{Listing}
\title{Pamba: Enhancing Global Interaction in Point Clouds via State Space Model}
\author{
   Zhuoyuan Li\textsuperscript{\rm 1}\equalcontrib,  Yubo Ai\textsuperscript{\rm 1}\equalcontrib, Jiahao Lu\textsuperscript{\rm 1},\; ChuXin Wang\textsuperscript{\rm 1}, Jiacheng Deng\textsuperscript{\rm 1}, \\Hanzhi Chang\textsuperscript{\rm 1}, Yanzhe Liang\textsuperscript{\rm 1}, Wenfei Yang\textsuperscript{\rm 1}, Shifeng Zhang\textsuperscript{\rm 2}, Tianzhu Zhang\textsuperscript{\rm 1}\thanks{Corresponding author: Tianzhu Zhang.}
}
\affiliations{
      \textsuperscript{\rm 1}Deep Space Exploration Laboratory/School of Information Science and Technology,\\ University of Science and Technology of China \\ \textsuperscript{\rm 2}Sangfor Technologies Inc. \\
}

\usepackage{bibentry}

\def\name{Pamba}
\usepackage{subcaption}
\usepackage{amsmath}
\usepackage{amssymb}
\usepackage{booktabs} 
\usepackage{multicol}
\usepackage{multirow}

\usepackage[utf8]{inputenc} 
\usepackage[T1]{fontenc}    
\usepackage{url}            
\usepackage{booktabs}       
\usepackage{amsfonts}       
\usepackage{nicefrac}       
\usepackage{microtype}      
\usepackage{xcolor}         
\usepackage{amsmath}
\usepackage{algorithmic}
\usepackage{algorithm}
\usepackage{amssymb}
\usepackage{multirow}
\usepackage{subcaption}
\usepackage{makecell}
\usepackage{multicol}
\usepackage{enumitem}
\usepackage{xcolor}
\usepackage{bm}
\usepackage{caption}

\begin{document}

\maketitle

\begin{abstract}
Transformers have demonstrated impressive results for 3D point cloud semantic segmentation. However, the quadratic complexity of transformer makes computation costs high, limiting the number of points that can be processed simultaneously and impeding the modeling of long-range dependencies between objects in a single scene. Drawing inspiration from the great potential of recent state space models (SSM) for long sequence modeling, we introduce Mamba, an SSM-based architecture, to the point cloud domain and propose \name{}, a novel architecture with strong global modeling capability under linear complexity. Specifically, to make the disorderness of point clouds fit in with the causal nature of Mamba, we propose a multi-path serialization strategy applicable to point clouds. Besides, we propose the ConvMamba block to compensate for the shortcomings of Mamba in modeling local geometries and in unidirectional modeling. \name{} obtains state-of-the-art results on several 3D point cloud segmentation tasks, including ScanNet v2, ScanNet200, S3DIS and nuScenes, while its effectiveness is validated by extensive experiments.
\end{abstract}

%

\section{Introduction}
3D point cloud semantic segmentation is a fundamental task in 3D scene understanding, which aims to predict the semantic labels for all points in the scene. As a critical technique for understanding realistic scenes, 3D point cloud semantic segmentation has various applications, including robotics \cite{robot}, automatic driving \cite{KittiDataset, nuscenes, nuscenes_lidar, waymo} and AR/VR \cite{VR}. However, the interaction between different points at different scales in the scene poses challenges to precise 3D point cloud semantic segmentation. 

\begin{figure}[t]
\centering
\includegraphics[width=0.85\columnwidth]{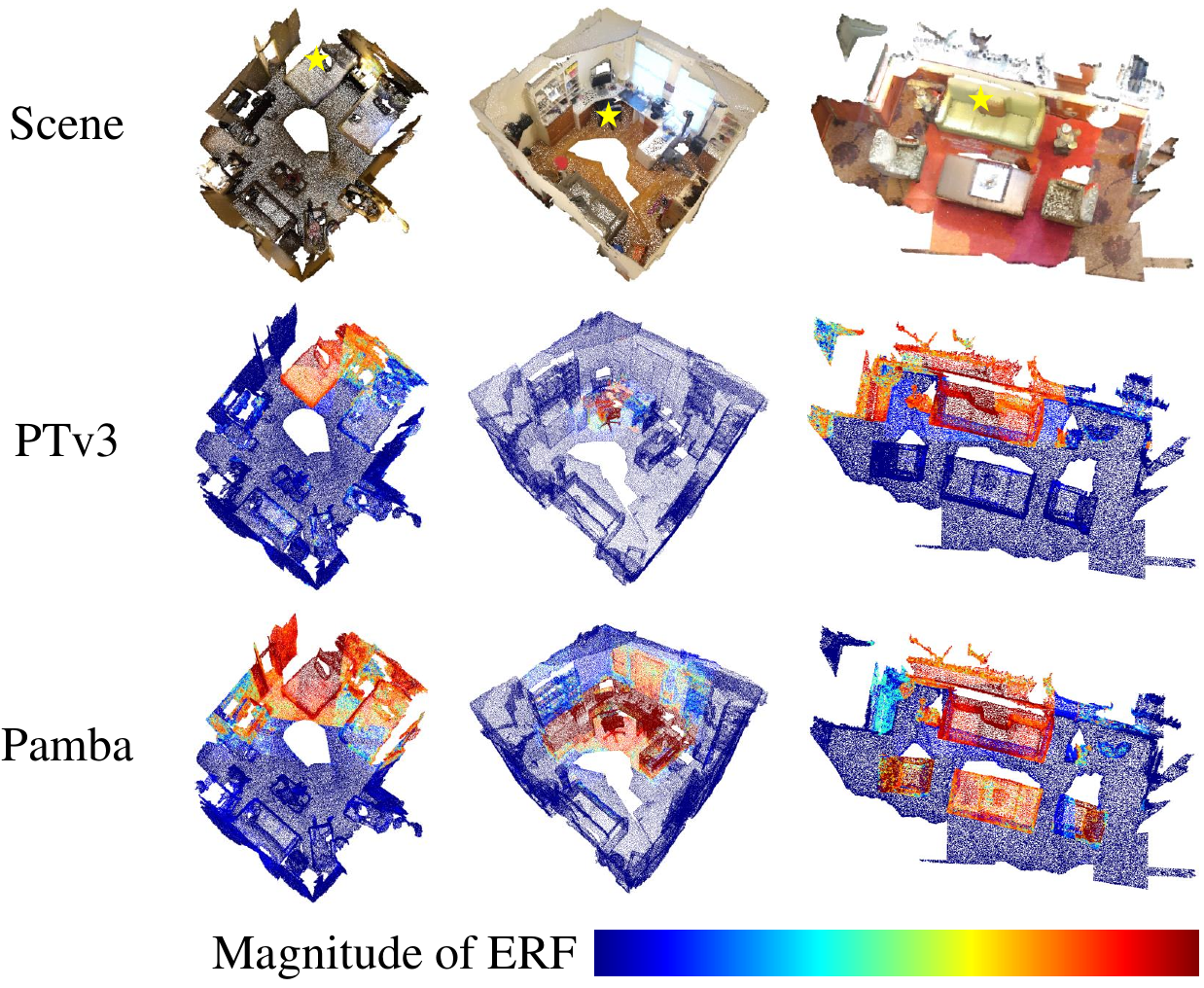}
\caption{Visualization of effective receptive fields (ERF) of the point of interest on ScanNet200 dataset. The yellow star represents the position of the point of interest. \name{} shows larger ERF and the ability of handling long-range interactions between different objects in a scene. More illustrations are provided in Appendix.}
\label{figure:pacman}
\end{figure}

To overcome the above challenges, a variety of 3D semantic segmentation methods have been proposed, which mainly fall into two categories: voxel-based methods and point-based methods. Voxel-based methods first quantize irregular point clouds into regular voxel representations and then perform 3D convolutions on the voxels \cite{voxnet, voxelrcnn}. The cubic growth in the number of voxels as a function of resolution leads to significant inefficiency, which is not solved until the proposal of sparse convolutions \cite{sparseconv1, sparseconv2}. However, the quantization loss during voxelization always exists. Therefore, some point-based methods that directly handle the points are proposed \cite{pointnet, pointnet++, kpconv, dgcnn}. The pioneering work PointNet \cite{pointnet} adopts permutation-invariant operators to aggregate features across the whole point cloud. Recently, inspired by the big success of transformer in the field of vision \cite{ViT, cvt, swin} and natural language processing \cite{bert, chatgpt1, llama}, many works \cite{PointTransformerv1,PointTransformerv2,octformer, flatformer,sphericalformer,swin3d}, which are categorized into point-based methods, incorporate transformer into point cloud analysis and achieve exceptional performance. Point Transformer \cite{PointTransformerv1} utilizes KNN \cite{knn} to construct the neighbourhood, in which local attention is performed. PTv2 \cite{PointTransformerv2} adopts grids to partition the point cloud into non-overlapping patches and performs attention mechanism per patch. Further, FlatFormer \cite{flatformer}, OctFormer \cite{octformer}, and PTv3 \cite{PointTransformerv3} adopt serialization-based methods to partition point clouds. The superior performance of transformer-based methods is attributed to transformer's strong ability of modeling long-range dependencies in large reception fields \cite{attentionisallyouneed, dotransformerseelikecnn}. However, transformer-based approaches are flawed in terms of scalability. Specifically, the quadratic complexity of transformer makes computation costs high, limiting the number of points that can be processed simultaneously and impeding the modeling of long-range interactions between objects in a single scene. For ease of illustration, we visualize the effective receptive fields (ERF) of a previous method, PTv3, in Fig. 1. The visualization demonstrates that previous methods focus more on semantically consistent neighbouring points but fail to capture interactions between objects due to limited ERF. As an example, in this context, a bed is more likely to appear against a wall and another bed, a desk is more likely to come in a set with a chair, a sofa is more likely to have a table and other sofas next to it. This prior knowledge matches human visual perception and can make models perform better.

Recent research advancements have sparked considerable interest in state space models (SSM) \cite{ssm, lssl, dss, s4}, which excel at capturing long-range dependencies under linear complexity while benefiting from parallel training . In particular, an SSM-based architecture, Mamba, demonstrates superior performance for NLP tasks to rival transformer \cite{mamba}. This leads us to think: Is it possible to introduce Mamba into the point cloud scene understanding tasks to solve the scalability problem of existing transformer-based methods? However, we find the direct application of Mamba into 3D scene understanding tasks results in poor performance. After analysis, we point out three main problems with processing point clouds using Mamba. 
\textbf{1)} Permutation sensitivity: Mamba is designed to process the causal sequence \cite{mamba}, which is highly sensitive to the input order. Different orders of input points can result in different outputs and greatly impact the final result.
\textbf{2)} Insufficiently strong local modeling ability: Mamba enhances the modeling of global features by compressing all contexts into a specific state \cite{mamba}. However, many irrelevant contexts are redundant for local modeling, which sacrifices the representation quality of local geometries.   
\textbf{3)} Unidirectional modeling: Mamba performs unidirectional modeling. For a point cloud sequence processed with Mamba, a point can only interact with points before this point, but not the points after this point, which hinders the bidirectional interaction between different points.

Based on the above analyses, we propose a novel point cloud scene understanding framework, \textbf{\name{}}, to address the above problems and fully unleash the potential of Mamba in the point cloud domain. First, we propose the \textbf{multi-path serialization} strategy to adapt to the permutation sensitivity of Mamba. Specifically, it rearranges the unordered point cloud to an ordered point sequence according to a pre-defined pattern so that points that are adjacent in sequence are also neighbouring in space. Based on previous Hilbert and z-order serialization patterns, we redesign a new pattern, named hz serialization, for a better fusion of spatial information in different perspectives within each layer. Then, we randomly assign hz serialization patterns to different blocks to enable the interaction of spatial information in different perspectives across layers.
Besides, we propose the \textbf{ConvMamba} block to compensate for the shortcomings of Mamba in modeling local geometries and in unidirectional modeling. In detail, it combines convolutions with Mamba to extract both long-distance dependencies and local geometries simultaneously. Moreover, bidirectionality is introduced to ConvMamba to enhance the bidirectional interaction between points.
Beyond the above, to enable global modeling, \name{} processes the entire point cloud directly, unlike previous transformer-based methods \cite{PointTransformerv1,PointTransformerv2,octformer, flatformer,sphericalformer,swin3d}, which split the point cloud into patches and then process them separately.
Taking ScanNet v2 (the average number of points is 148k) as an example, \name{} directly processes over 100k points without dividing the points into groups.

As shown in Fig. 1, \name{} has a larger ERF and can capture interactions not only between semantically similar neighbouring parts, but also between objects, e.g., the interaction between sofa and table or the interaction between desk and chair. Notably, \name{} focuses on how to utilize Mamba to capture long-range interactions between points that are hindered in transformer, rather than on some intricate design. The contributions can be summarised as follows:
\begin{itemize}
  \item    
    We propose a new framework \name{} as a direct application of Mamba to semantic segmentation of 3D point clouds, which captures long-range interaction under linear complexity.
  \item 
    We propose the multi-path serialization strategy and the ConvMamba block to help Mamba better adapted to point clouds. The former one enables the model to capture spatial information in different perspectives while the latter one compensates for the shortcomings of Mamba in modeling local geometries and in unidirectional modeling.
  \item 
    We conduct extensive experiments and ablation studies to validate our design choices. \name{} achieves state-of-the-art performance on several highly competitive point cloud segmentation tasks, including ScanNet v2, ScanNet200, S3DIS, and nuScenes.
\end{itemize}

\begin{figure*}[t]
  \centering
  \includegraphics[scale=0.65]{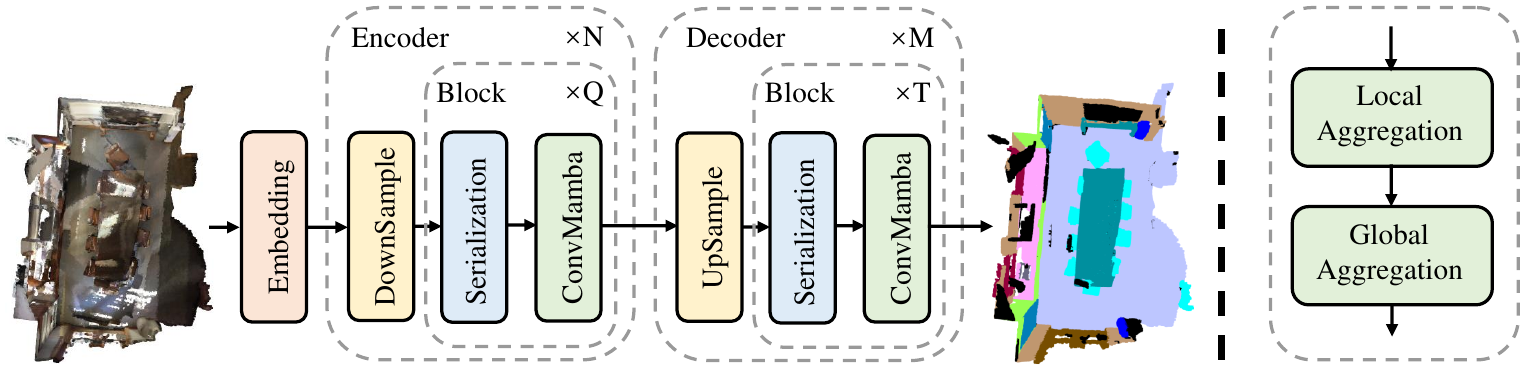}
  \caption{Left: The overall architecture of \name{}; Right: ConvMamba block.}
  \label{fig:overall}
\end{figure*}

\paragraph{Point cloud transformer.}
It is natural to extend transformer into point cloud understanding after the big success of vision transformers \cite{ViT}, which can be counted as a sub-category of point-based methods. PCT \cite{PCTguo} and Point Transformer \cite{PointTransformerv1} are the pioneers in introducing transformer into the field of point cloud. PCT \cite{PCTguo} directly applies global attention to all points inside the point cloud and thus can only handle point clouds with a few thousand points due to the quadratic complexity of transformer. In contrast, Point Transformer \cite{PointTransformerv1} first extracts the points' neighbourhood by KNN \cite{knn}, in which the local attention is then applied, achieving much less memory costs than PCT. Following Point Transformer \cite{PointTransformerv1}, many transformer-based methods spring up and achieve state-of-of-the-art performance, such as PTv2 \cite{PointTransformerv2}, PTv3 \cite{PointTransformerv3}, Stratified Transformer \cite{stratifiedformer}, PatchFormer \cite{patchformer}, QueryFormer \cite{lu2023query} etc.

\paragraph{Point cloud serialization.}
Unlike the traditional paradigm of partitioning the point cloud into patches through KNN or similar methods,  several recent works \cite{flatformer, octformer, PointTransformerv3} propose to arrange the irregular and scattered points into a structured point sequence by directly sorting the whole point cloud through specific patterns, which are categorized as serialization-based methods. These approaches achieve strong performance while keeping memory overhead low, as sorting algorithms could be much more efficient than KNN-related algorithms. FlatFormer \cite{flatformer} utilizes windows to partition the point cloud, then sorts windows along $x$ and $y$ axis. However, directly sorting along axes may result in the destruction of spatial proximity, leading to a performance drop. OctFormer \cite{octformer} follows the octree structure to sort the whole point cloud. The nature of the octree ensures that the spatial proximity of the unstructured point cloud can be well preserved during the serialization. Based on prior works, PTv3 \cite{PointTransformerv3} further combines the Hilbert serialization and the z-order serialization, unleashing the greater potential of serialization-based methods.

\paragraph{State space models.}
State space models (SSM) originate from the classic Kalman filter model in control systems. SSM can either model long-range interactions like RNN or be trained in parallel like transformer, achieving high efficiency. Recently, many variants of SSM have been proposed, including linear state-space layers \cite{lssl}, structured state space model \cite{s4}, and diagonal state space \cite{dss}. Mamba \cite{mamba} is the state-of-the-art SSM-based architecture. It proposes a selective mechanism so that the model parameters vary with inputs, allowing the model to compress context selectively according to current input \cite{mamba}. This principle further enhances the ability to model long-range dependencies. Several recent works adapt Mamba to different fields, including vision \cite{vmamba1, vmamba2, Swin-UMamba, mambair, efficientvmamba}, graph neural network \cite{graphmamba1, graphmamba2, graphmamba3} and video \cite{videomamba1, videomamba2}. 
Some concurrent works PM \cite{pointmambabaixiang} PoinTramba \cite{pointramba}, SPM \cite{serializedpm} and PCM \cite{pointmambayanshuic} also apply Mamba to 3D point clouds. However, PM is a pre-training architecture similar to Point-MAE \cite{pointmae}. PoinTramba enhances PM by combining attention with Mamba. PCM and SPM only apply Mamba within the neighbourhood extracted by KNN, which ignores the global interaction and does not fully utilize Mamba's ability to capture long-range dependencies. Besides, the above works are mainly evaluated on object-level datasets \cite{modelnet40, shapenet, scanobjectnn} and not on scene-level datasets, as the latter one is crucial for 3D scene understanding.

\section{Method}

The general structure of Pamba is shown in Fig. 2. Pamba is inherited from PTv3 \cite{PointTransformerv3}, the state-of-the-art backbone for point cloud perception tasks. PTv3 is chosen for its concise pipeline as well as its strong performance.  
We start with a brief preliminary of state space models and space-filling curves in Section 3.1. In Section 3.2, we introduce the multi-path serialization strategy. In Section 3.3, we introduce ConvMamba, the main block of \name{}.

\subsection{Preliminary}
\label{sec:method:preliminary}
\paragraph{State space model.}
The state space model (SSM) is initially introduced in the field of control engineering to model dynamic systems. Specifically, the SSM in deep learning encompasses three key variables: the input sequence \(x(t)\), the latent state representation \(h(t)\), and the output sequence \(y(t)\). Additionally, it includes two fundamental equations: the state equation and the observation equation, with $\textit{\textbf{A}}$, $\textit{\textbf{B}}$ and $\textit{\textbf{C}}$ being system parameters. The SSM is formulated in Eq. \ref{eq:ssm}.

\begin{equation}
\begin{aligned}
\label{eq:ssm}
h'(t) &= \textit{\textbf{A}}h(t)+\textit{\textbf{B}}x(t), \\
y(t) &= \textit{\textbf{C}}h(t).
\end{aligned}
\end{equation}

\begin{figure*}[t]
  \centering
  \includegraphics[scale=0.55]{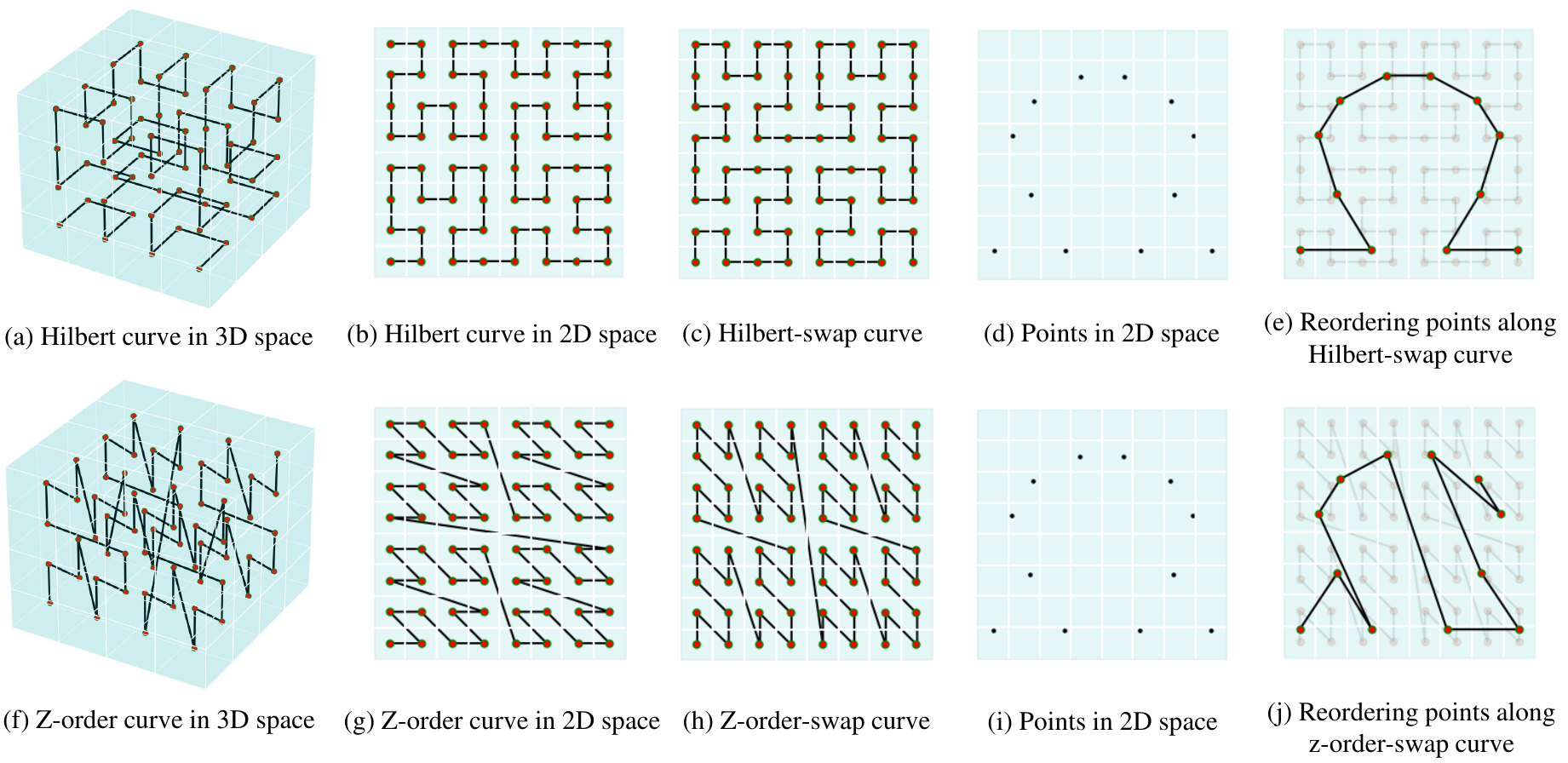}
  \caption{Space-filling curves.}
  \label{fig:serialization}
\end{figure*}

\paragraph{Mamba.}
Discretizing the SSM is crucial because it is initially designed for continuous systems and cannot handle discrete data such as images or point clouds. \citeauthor{mamba} utilizes the zero-order hold technique to discretize the SSM with a time step $\boldsymbol{\varDelta}$. In detail, the continuous parameters $\textit{\textbf{A}}$, $\textit{\textbf{B}}$, $\textit{\textbf{C}}$ are transformed to discrete parameters $\overline{\textit{\textbf{A}}}$, $\overline{\textit{\textbf{B}}}$, $\overline{\textit{\textbf{C}}}$ as shown in Eq. \ref{eq:s4}

\begin{align}
    \overline{\textit{\textbf{A}}}=e^{\boldsymbol{\varDelta}\textit{\textbf{A}}},\quad
    \overline{\textit{\textbf{B}}}=(e^{\boldsymbol{\varDelta} \textit{\textbf{A}}}-\textit{\textbf{I}})\textbf{\textit{A}}^{-1}\textit{\textbf{B}},\quad
    \overline{\textit{\textbf{C}}}=\textit{\textbf{C}}.
    \label{eq:s4}
\end{align}

After discretization, the calculation process of SSM can be simplified into a convolution operation, enabling the entire SSM to be trained in parallel, as shown in Eq. \ref{eq:cnn}.
\begin{align}
   \overline{\textbf{y}}=\textbf{x}\circledast \overline{\textit{\textbf{K}}},\quad 
    \quad \overline{\textit{\textbf{K}}}=(\textit{\textbf{C}}\overline{\textit{\textbf{B}}},\textit{\textbf{C}}\overline{\textit{\textbf{AB}}},\ldots,\textit{\textbf{C}} \overline{\textit{\textbf{A}}}^{k-1}\overline{\textit{\textbf{B}}}).
    \label{eq:cnn}
\end{align}

\paragraph{Space-filling curve.}

A space-filling curve \cite{spacefillingcurve} is a curve that fills a multi-dimensional space. When the dimension equals three in the context of point clouds, the space-filling curve traverses all points within a discrete 3D cube without repetition. The best-known space-filling curves include Hilbert curve \cite{hilbert} and z-order curve \cite{z-order}, as shown in Fig. \ref{fig:serialization} (a), (b), (f) and (g) with a dimension of three and two respectively. Z-order curves are known for its high efficiency, while Hilbert curves are known for its locality-preserving property.
For ease of illustration, we elaborate the following in 2D.

The shown space-filling curves in Fig. \ref{fig:serialization} (b) and (g) adopt a traversal along $x$, $y$, and $z$ axes in the order of priority. By simply changing the order of the three axes, similar variants of the space-filling curves are obtained. Here we show two other variants of the space-filling curve named Hilbert-swap curve (swap stands for swapping $x$ axis and $y$ axis) and z-order-swap curve by exchanging the order of the $x$ axis and $y$ axis for traversal as shown in Fig. \ref{fig:serialization} (c) and (h).

Based on the space-filling curve, an intuitive idea is that points in space can be sorted into a 1-D point sequence along the space-filling curve, which we call point cloud serialization. 
OctFormer \cite{octformer} and PTv3 \cite{PointTransformerv3} are the pioneers that apply the above curves to point clouds. OctFormer \cite{octformer} utilizes z-order curves for serialization, while PTv3 utilizes both Hilbert curves and z-order curves.
The spatial proximity in point clouds can be preserved well through point cloud serialization, i.e., points that are adjacent in sequences are also neighbouring in point clouds. An example of some randomly located points is shown in Fig. \ref{fig:serialization} (d) and (i), which are separately sorted using the Hilbert-swap curve and the z-order-swap curve in Fig. \ref{fig:serialization} (e) and (j). It could be observed that the Hilbert curve preserves better spatial proximity than the z-order curve, which is already proved by some previous work \cite{comparehilbert&z}.

\subsection{Multi-path serialization strategy}

As Hilbert-based serialization keeps better spatial proximity than z-order-based serialization \cite{comparehilbert&z}, it is natural to apply Hilbert-based serialization rather than z-order serialization. However, we observe poor performance when simply applying Hilbert-based serialization in each ConvMamba block in Tab. \ref{Tab:serial}. We attribute this to a single ordering pattern lacking the spatial relationships in multiple perspectives provided by other ordering patterns. 
PTv3 \cite{PointTransformerv3} solves this by proposing the "shuffle order strategy" that assigns different serialization patterns to different layers so that the spatial relationships in multiple perspectives provided by different serialization patterns are mixed across layers. However, within each layer, the spatial information therein is still obtained from a single serialization pattern.

\begin{figure}[t] 
\centering 
\includegraphics[width=1\columnwidth]{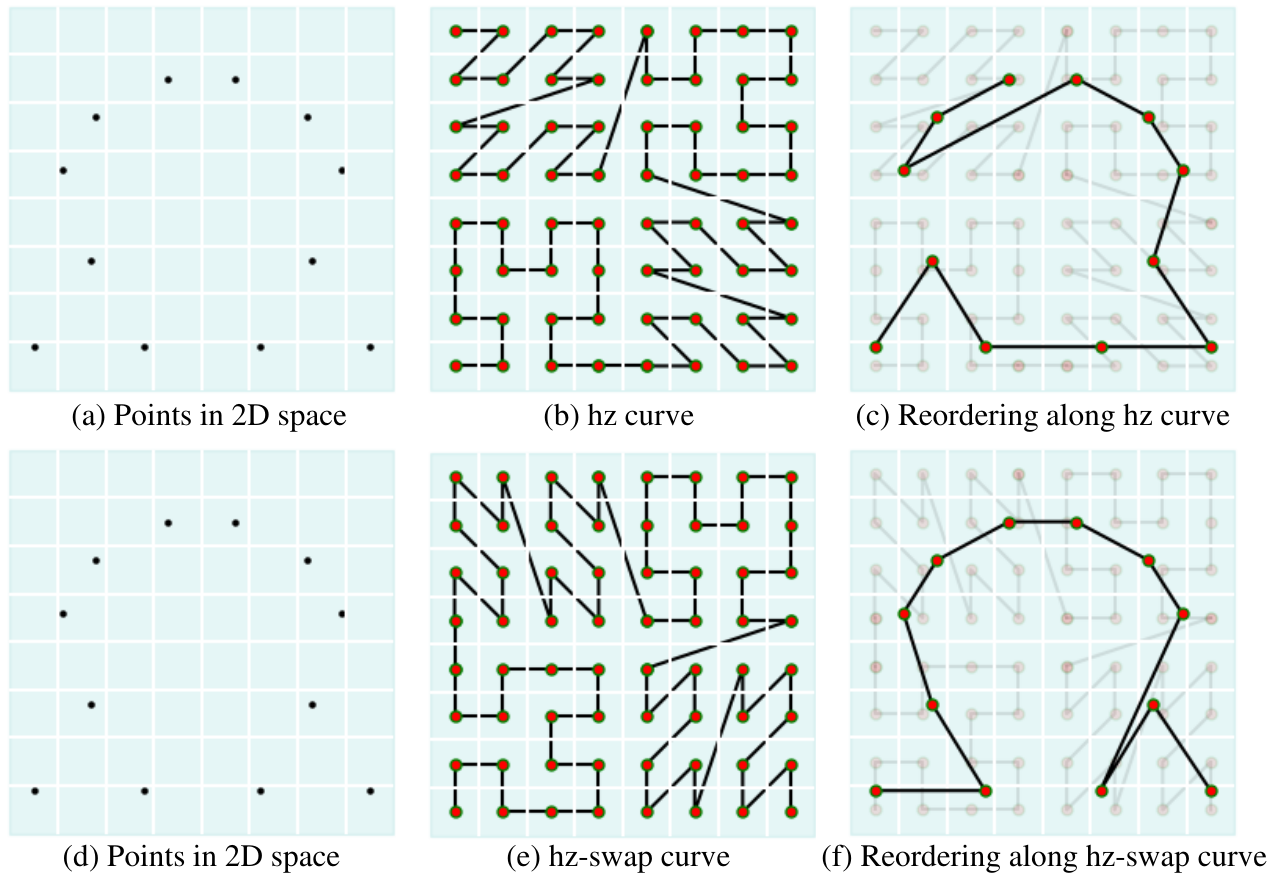} 
\caption{Our introduced hz curve and hz-swap curve.}
\label{fig:hz}
\end{figure}

To further enhance the fusion of multi-perspective spatial relationships, we propose the multi-path serialization strategy. In detail, we redesign two brand new serialization patterns, namely the hz curve and the hz-swap curve, by integrating both Hilbert curves and z-order curves as shown in Fig. \ref{fig:hz}. Examples of sorting along the hz curve and the hz-swap curve are shown in Fig. \ref{fig:hz} (c) and (f). Then, for each ConvMamba block, we randomly allocate one pattern from the hz curve and the hz-swap curve to the block. With this strategy, the model captures multi-perspective spatial relationships in each layer and achieves better robustness.

\subsection{ConvMamba}
\label{sec:method:convmamba}
In this section, we introduce the ConvMamba block, which aims to synergistically capture global dependencies and local features. It consists of two stages, local aggregation and global aggregation, going serially as shown in Fig. \ref{fig:overall}. 

\begin{figure}[t] 
\centering 
\includegraphics[width=1\columnwidth]{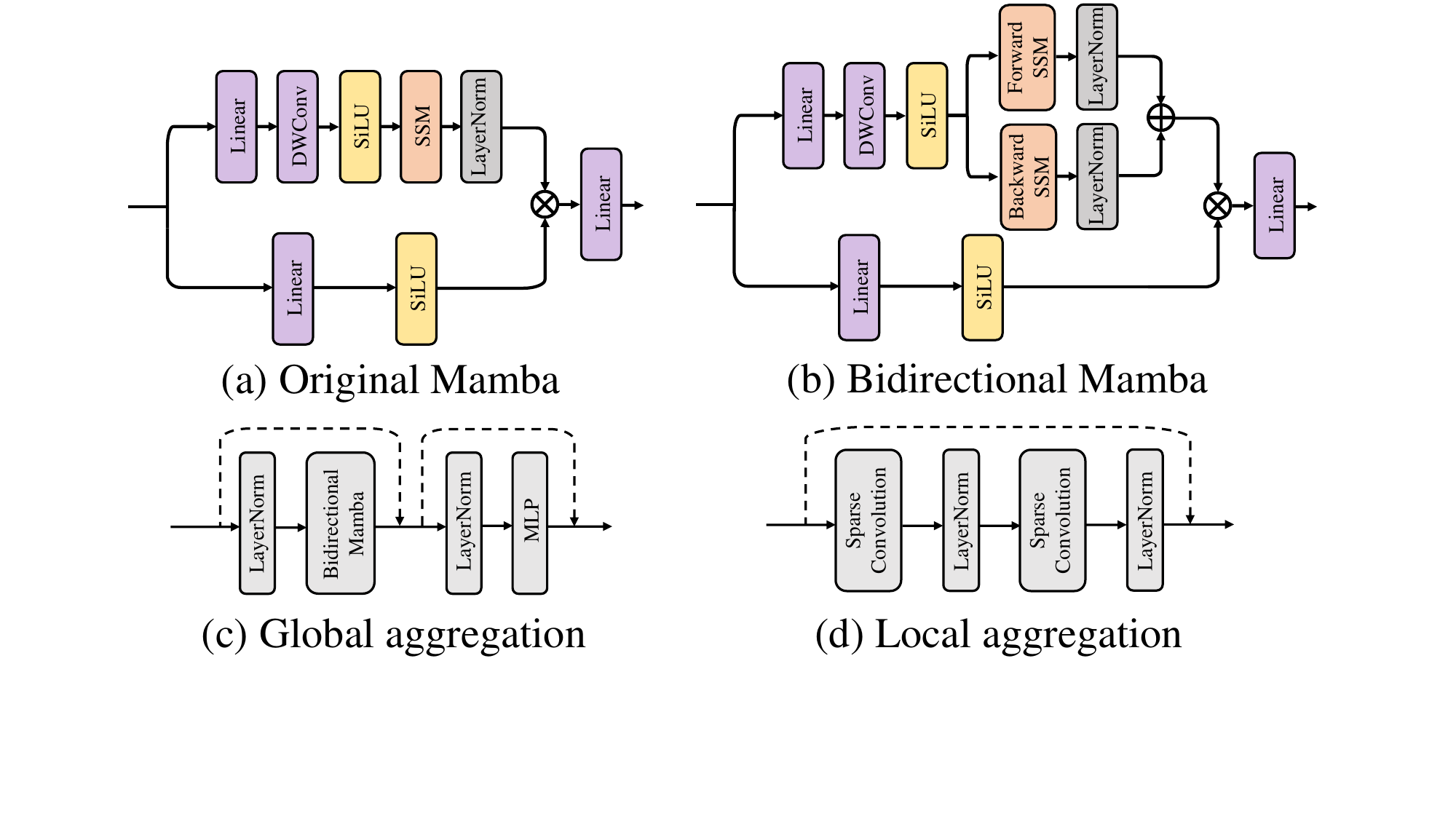} 
\caption{Bidirectional Mamba. (a) Original Mamba structure; (b) Proposed bidirectional Mamba; (c) Global aggregation.}
\label{fig:bimamba}
\end{figure}

\paragraph{Global Aggregation.}
Due to the linear complexity of Mamba, we apply Mamba to process the whole point cloud (more than 100k points) at once instead of applying Mamba within patches like what point transformers \cite{PointTransformerv1, PointTransformerv2} does, realizing long-range interaction. However, one inherent drawback in the original Mamba \cite{mamba} is its causality as shown in Fig. \ref{fig:bimamba} (a), meaning a point in the serialized point sequence can only interact with points before this point, not the points after this point. In other words, the block in Fig. \ref{fig:bimamba} (a) only scans the input sequence in one direction. To address this, we introduce the bidirectional mamba mechanism as shown in Fig. \ref{fig:bimamba} (b). 
Specifically, the whole point cloud is scanned from both forward and backward directions, enabling each point capable of interacting with points on either side of it. It is worth noting that `forward SSM' and `backward SSM' share the same parameters, which is different from some other Mamba-based works \cite{vmamba1, pointmambayanshuic}. This principle corresponds with our intention of obtaining consistent features from two opposite scans, putting a consistency constraint on both opposite scans.
We then follow the traditional transformer block \cite{attentionisallyouneed} and pre-norm pattern \cite{prenorm} to construct the global aggregation stage by applying an MLP after bidirectional Mamba, both with normalization and skip connection as shown in Fig.\ref{fig:bimamba} (c).

\paragraph{Local Aggregation.}
Bidirectional Mamba is applied to the whole point cloud to capture long-range dependencies by compressing all context into a hidden state. However, some irrelevant contexts are redundant for local modeling, which sacrifices the representation quality of local geometries. Unfortunately, local features are proven to be essential to point clouds \cite{kpconv, pointvector, condaformer}. To address this, we add a local aggregation stage right before the global aggregation stage to explicitly aggregate local information and compensate for Mamba's shortcomings in modelling local geometries. 
Inspired by \cite{cpe,octformer,PointTransformerv3}, which verified that various convolutions not only aggregate local features but also provide location information, we simply utilize sparse submanifold convolutions to form the local aggregation stage for its high efficiency and low memory usage.

Specifically, we adopt two consecutive sparse convolutions to achieve a larger receptive field than a single convolution, as shown in Fig. \ref{fig:bimamba} (d), as the local information extracted from a larger context is more helpful to the long-range interaction of over 100K points, which is verified in ablations.

\section{Experiments}
In this section, we aim to evaluate the effectiveness of our proposed \name{}. We introduce the main results of 3D semantic segmentation tasks in section 4.1. In section 4.2, we evaluate the efficiency of \name{}. In section 4.3, we conduct ablation studies on the design choice of \name{}. In section 4.4, we discuss limitations and future work.

\subsection{Semantic segmentation}
\label{sec:exp:main_result}

\paragraph{Dataset.}
We evaluate \name{} on four datasets: ScanNet v2\cite{scannet}, ScanNet200 \cite{scannet200}, S3DIS \cite{s3dis} and nuScenes \cite{nuscenes, nuscenes_lidar}. ScanNet v2 is a commonly used indoor dataset, containing 1513 room scans in total, the average point number of which is 148k. ScanNet200 shares the same data with ScanNet but has 200 semantic categories, making precise predictions more difficult.
The S3DIS dataset is another indoor dataset consisting of 271 rooms in six areas from three different buildings, with a total of 13 semantic categories.
nuScenes is an outdoor dataset containing 1,000 driving sequences in total, which is usually more difficult to handle than indoor datasets.
ScanNet, ScanNet200 and nuScenes follow the official data splits to generate corresponding training set, validation set and test set. For S3DIS, the data is split into six areas, the fifth of which is withheld during training and used for evaluation. 

\paragraph{Setting.}
We train our model on 4 RTX 3090 GPUs. AdamW \cite{adamw} is adopted for parameter optimization. ScanNet and ScanNet200 are trained 800 epochs, S3DIS is trained 3000 epochs and nuScenes is trained 50 epochs. The sum of CrossEntropy loss and Lovasz loss \cite{lovasz} is adopted as the overall loss for \name{} as shown in Eq. \ref{eq:loss}, where $L_{CE}$ represents CrossEntropy loss and $L_{L}$ represents Lovasz loss. $\lambda_1$ and $\lambda_2$ are both empirically set to 0.5 during implementation.
\begin{align}
    L = \lambda_1 \cdot L_{CE} + \lambda_2 \cdot L_{L}
    \label{eq:loss}
\end{align}

\begin{table}[t]
  \setlength{\tabcolsep}{0.9mm}
\begin{tabular}{lcccc}
\toprule
\multicolumn{1}{c}{\multirow{2}{*}{Methods}}             &  Scannet       &  ScanNet200    &  nuScenes      & S3DIS\\
\multicolumn{1}{c}{}                                     &  Val           &  Val           &  Val           & Area 5\\
\midrule
\textbf{Point} \\ 
\midrule
 PointNet++             &  53.5          & -             & -           & -  \\
 PointConv              &  61.0          & -             & -           & -  \\
 KPConv                     &  69.2          & -             & -           & 67.1  \\
 PointNeXt               &  71.5          & -             & -           & 70.5  \\
 PointMetaBase       &  72.8          & -             & -           & 71.3  \\ 
 \midrule
\textbf{Transformer} \\ 
 \midrule
 PTv1           &  70.6          &  27.8          & -          & 70.4   \\
 PTv2           &  75.4          &  30.2          &  80.2      & 71.6    \\
 StratifiedFormer &  74.3          & -             & -           & 72.0  \\
 OctFromer               &  75.7          &  32.6          & -          & -   \\
 SphereFormer      & -             & -             &  78.4        & -  \\
 Swim3D                     &  76.4          & -             & -           & 72.5  \\ 
 PTv3                     &77.5          & 35.2             & 80.4       & 73.4   \\
 \midrule
\textbf{Mamba} \\ 
\midrule
PCM     &-   &-   &-   &63.4 \\
\textbf{\name{}}          &\textbf{77.6} &\textbf{36.3} &\textbf{80.4} &\textbf{73.5} \\ 
\bottomrule
\end{tabular}
\caption{Semantic segmentation result}
\label{Tab:main_result}
\end{table}

\paragraph{Main results.}
We compare \name{} with a variety of previous state-of-the-art models, including PointNet++ \shortcite{pointnet++}, PointConv \shortcite{pointconv}, KPConv \shortcite{kpconv}, Cylender3D \shortcite{cylinder3d}, PointNeXt \shortcite{pointnext}, PointMetaBase \shortcite{pointmetabase}, PTv1 \shortcite{PointTransformerv1}, PTv2 \shortcite{PointTransformerv2}, StratifiedFormer \shortcite{stratifiedformer}, OctFormer \shortcite{octformer}, SphereFormer \shortcite{sphericalformer}, Swim3D \shortcite{swin3d}, PTv3 \shortcite{PointTransformerv3} and PCM \shortcite{pointmambayanshuic}, using mean Intersection over Union (mIoU) as the metric in Tab. \ref{Tab:main_result}. 
\name{} shows significant priority and exceeds most previous methods. Additionally, \name{} demonstrates competitive performance with PTv3 \shortcite{PointTransformerv3}, the most advanced transformer-based method, and even surpasses it on some datasets.

In detail, for indoor datasets, on ScanNet v2 and S3DIS, \name{} shows comparable performance with PTv3, achieving a mIOU of 77.6\% and 73.5\% respectively. For outdoor scenes, \name{} is also capable of reaching a mIOU of 80.4\% on nuScenes, showing competitive performance and demonstrating \name{}'s strong generalization ability over both indoor scenes and outdoor scenes.

When it comes to complex scenes with more categories on ScanNet200, \name{} further demonstrates superior performance, achieving a mIOU of 36.3\% and surpassing the previous state-of-the-art by 1.1\%, which indicates \name{}'s ability of handling complexly-labelled scenarios like real-life scenes. Besides, we conclude that the long-range interaction constructed by \name{} is more effective in complex scenarios with more categories than in simple scenarios. In simple scenarios with dozens of categories, modelling local geometries is enough to predict the right label and interactions between objects of different categories may help less. In complex scenarios with hundreds of categories, complicated coupling between points of different categories could lead to less accurate local modelling. In this case, the constructed interactions by \name{} can effectively alleviate the complexity that comes from excessive categories and enhance model performance.

\begin{table}[t]
  \centering
  \setlength{\tabcolsep}{1mm}
  \begin{tabular}{llcccc}
    \toprule
    \multicolumn{1}{l}{\multirow{2}{*}{Methods}}& \multicolumn{2}{c}{\textbf{Training}} & \multicolumn{2}{c}{\textbf{Inference}}\\
    
    \cmidrule(r){2-3} \cmidrule(r){4-5}
         &  Latency &  Memory &  Latency &  Memory
    \\
    \midrule
     OctFormer \shortcite{octformer}     &  357ms     &  9.5G     &  120ms      &  9.3G     \\
     Swin3D \shortcite{swin3d}     &  758ms     &  10.3G    &  529ms     &  7.0G     \\
     PTv2 \shortcite{PointTransformerv2}     &  398ms     &  10.1G    &  230ms     &  13.0G     \\
     PTv3 \shortcite{PointTransformerv3}     &  \textbf{223ms}     &  5.2G    &  \textbf{108ms}     &  \textbf{4.5G}     \\
     \midrule
    \textbf{\name{}} 
    & {249ms} & \textbf{5.2G} 
    & {179ms} & {4.8G} \\
    \bottomrule
  \end{tabular}
  \caption{Model Efficiency}
  \label{Tab:efficiency}
\end{table}

\subsection{Model efficiency}
\label{sec:exp:efficiency}
We measure the efficiency of \name{} through two metrics: mean latency and mean memory consumption on ScanNet200 dataset. All measurements are taken on a single RTX 3090 and are compared with previous methods as shown in Tab. \ref{Tab:efficiency}. Specifically, mean memory consumption is the memory per GPU recorded during training divided by the batch size. 

\paragraph{Memory consumption.}
Tab. 2 compares the mean memory consumption of \name{} with previous works during both the training stage and inference stage. During training, \name{} demonstrates a low memory consumption compared with all previous work. During inference, \name{} exhibits a slightly larger memory consumption, 0.3G greater than PTv3, but is still lower than other methods.

\paragraph{Model latency.}
\name{} maintains a better latency than many previous methods, while is still slightly slower than some efficient transformer-based models, such as OctFormer \cite{octformer} and PTv3, as shown in Tab. \ref{Tab:efficiency}.  We attribute this phenomenon to the poor parallelism of our proposed \name{}. 
In detail, OctFormer and PTv3 follow the common paradigm that first partitions the whole point cloud into patches with the same number of points, which can then be handled by self-attention mechanism in parallel at patch level. 
However, our \name{} handles the whole point cloud at once under the global reception field, with each point being calculated serially. This intrinsic difference is the key to the long-range interaction in \name{}, while inevitably making \name{} a bit slower. We posit this trade-off is beneficial overall.
Additionally, transformer-based methods benefit from FlashAttention \cite{flashattention}, an algorithm for deep optimization of attention mechanism in hardware, while SSMs have yet to be studied enough in terms of acceleration as an emerging model.

\begin{table}[t]
  \centering
  \setlength{\tabcolsep}{4mm}
  \begin{tabular}{lc}
    \toprule
    Type of Mamba               &Val\\
    \midrule
    {No Mamba}                &{26.1}\\
    {Unidirectional Mamba}                &{33.2}\\
    {Bidirectional Mamba w/ different params}                &{35.7}\\
    \midrule
    \textbf{Bidirectional Mamba}          &\textbf{36.3}\\
    \bottomrule
\end{tabular}
\caption{Effectiveness of Mamba.}
\label{Tab:effectivenessmamba}
\end{table}

\begin{table}[t]
  \centering
  \setlength{\tabcolsep}{16.5mm}
  \begin{tabular}{cc}
    \toprule
    Depth                            &Val\\
    \midrule
    0                                &31.9\\
    1                                &35.9\\
    \textbf{2}                       &\textbf{36.3}\\
    3                                &36.0\\
    4                                &35.4\\
    \bottomrule
    \end{tabular}
  \caption{Depth of local aggregation.}
\label{Tab:depth}
\end{table}

\begin{table}[t]
  \centering
  \setlength{\tabcolsep}{1mm}
\begin{tabular}{lc}
    \toprule
    Serialization combination & Val \\
    \midrule
    Hilbert                               & 34.4 \\
    Z                               &34.3 \\
    Hilbert + Hilbert-swap               &35.0 \\
    Z + Z-swap               &34.7 \\
    hz               &35.1 \\
    Hilbert + Hilbert-swap + Z + Z-swap &35.8\\
    Hilbert + Hilbert-swap + Z + Z-swap + hz + hz-swap               &35.8 \\
    \midrule
    \textbf{hz + hz-swap} &\textbf{36.3}\\
    \bottomrule
\end{tabular}
  \caption{Different serialization combination.}
  \label{Tab:serial}
\end{table}

\subsection{Ablation study}
\label{sec:exp:ablation}

Here, we verify the key design choices of \name{}. All ablation studies are conducted on ScanNet200 validation set.

\paragraph{Effectiveness of Mamba.}
We verify the effectiveness of Mamba in Tab. \ref{Tab:effectivenessmamba}. In detail, three experiments are conducted. First, we conduct experiments by removing all Mamba modules in \name{}, which results in a pure-convolution model. Second, we replace the bidirectional Mamba with unidirectional Mamba. Third, we make the `forward SSM' and `backward SSM' of bidirectional Mamba use different parameters.
The result shows that pure-convolution (No Mamba) underperforms, indicating that the long-range interaction constructed by Mamba is essential. The bidirectionality further enhances the long-range interaction and achieves better performance. Besides, the consistency constraint on the forward scan and backward scan is effective.

\paragraph{Depth of local aggregation.}
In Tab. \ref{Tab:depth}, we conduct experiments on the effects of different numbers of sparse convolutions employed in the local aggregation stage (including zero, meaning no local aggregation). Tab. \ref{Tab:depth} shows a non-negligible performance gap between \name{} and \name{} without sparse convolutions, indicating that the local aggregation is essential to \name{}.
Besides, the depth of 1, 2, 3 and 4 demonstrates similar performance on ScanNet200. Since sparse convolutions are highly efficient, we simply adopt the depth of 2.

\paragraph{Serialization strategy.}
In Tab. \ref{Tab:serial}, we verify the effectiveness of different combinations of serialization patterns while other settings remain the same as described in the method. First, simply adopting a single serialization pattern generates a base performance. Further results suggest that the increase in the number of serialization patterns obviously improves the model's performance, which corresponds with our viewpoint that different serialization patterns offer different perspectives on the spatial relationship in point clouds. Besides, the Hilbert-based patterns generally outperform z-order-based patterns.
Furthermore, by mixing Hilbert and z-order, the hz-based curve makes the performance even more powerful, up to a mIOU of 36.3\%, indicating the mixing of spatial relationships in multiple perspectives within layers is beneficial.

\subsection{Limitation}
\label{sec:exp:limitation}
Even though \name{} achieves a considerable latency compared with many previous methods, it is still slightly slower than some efficient transformer-based models, such as OctFormer \cite{octformer} and PTv3 \cite{PointTransformerv3}. We analyze that this efficiency gap comes from the poor parallelism of \name{}. Specifically, OctFormer and PTv3 follow the traditional paradigm that first divides the whole point cloud into groups with the same number of points, which can then be handled by self-attention mechanism in parallel at group level. In contrast, \name{} handles the whole point cloud at once under the global reception field, with each point being calculated serially. This intrinsic difference inevitably makes \name{} a bit slower, reinforcing the need for continued exploration of efficient SSM mechanisms with better parallelism.
Besides, \name{} is only evaluated on semantic segmentation tasks and may be extended to other tasks in the future, e.g., instance segmentation and object detection.

\section{Conclusion}
We explore the direct application of Mamba to semantic segmentation of 3D point clouds, which is a challenging and versatile task for evaluating different techniques.
Unlike previous point transformers, we do not follow the common paradigm that first partitions the whole point cloud into patches and then processes each patch in parallel. In contrast, we treat the whole point cloud as a single `patch' and pass it to a Mamba layer to enable global interaction. We make some non-trivial improvements to adapt Mamba to the field of point clouds, including the multi-path serialization strategy and the ConvMamba block. Our \name{} exceeds the state of the art on many point cloud semantic segmentation tasks.


\section*{Acknowledgments}
This work was supported by National Natural Science Foundation of China (62306294, 62394354), Anhui Provincial Natural Science Foundation 2308085QF222, Youth Innovation Promotion Association.

\bibliography{aaai25}

\end{document}